\pdfoutput=1

\documentclass[11pt]{article}

\usepackage[final]{acl}

\usepackage{times}
\usepackage{latexsym}

\usepackage[T1]{fontenc}

\usepackage[utf8]{inputenc}

\usepackage{microtype}

\usepackage{inconsolata}

\usepackage{graphicx}

\usepackage{lipsum} 
\usepackage{stfloats} 
\usepackage{times}
\usepackage{latexsym}
\usepackage{hyperref}
\usepackage{amsmath}
\usepackage{amssymb}
\usepackage{array}
\usepackage{makecell}
\usepackage[T1]{fontenc}
\usepackage{inputenc}

\usepackage[utf8]{inputenc}

\usepackage{microtype}

\usepackage{inconsolata}

\usepackage{graphicx}
\usepackage{booktabs}
\usepackage{hyperref}
\usepackage{tabularx}

\usepackage{fontawesome5}

\usepackage{booktabs, colortbl} 
\usepackage{multirow,makecell} 
\usepackage{multicol}
\usepackage{amsmath}
\usepackage{amssymb}
\usepackage{caption}
\usepackage{subcaption}
\usepackage{titlesec}
\usepackage{pgfplots}
\usepackage[normalem]{ulem} 
\usepackage{enumitem}
\usepackage{graphicx} 
\usepackage{hyperref}
\usepackage{svg}

\usepackage{enumitem}
\usepackage[splitrule]{footmisc}
\setlist[itemize]{noitemsep, nolistsep}
\usepackage{color, colortbl}
	
\definecolor{Gray}{gray}{0.9}

\title{Middle-Layer Representation Alignment for Cross-Lingual Transfer in Fine-Tuned LLMs}

\author{Danni Liu \phantom{\and} Jan Niehues \\
        Karlsruhe Institute of Technology, Germany \\
        \texttt{\{danni.liu, jan.niehues\}@kit.edu}} 

\begin{document}
\maketitle

\begin{abstract}
While large language models demonstrate remarkable capabilities at task-specific applications through fine-tuning, 
extending these benefits across diverse languages is essential for broad accessibility.
However, effective cross-lingual transfer is hindered by LLM performance gaps across languages and the scarcity of fine-tuning data in many languages.
Through analysis of LLM internal representations from over 1,000+ language pairs,
we discover that middle layers exhibit the strongest potential for cross-lingual alignment.
Building on this finding, 
we propose a middle-layer alignment objective integrated into task-specific training.
Our experiments on slot filling, machine translation, and structured text generation show consistent improvements in cross-lingual transfer, especially to lower-resource languages.
The method is robust to the choice of alignment languages and generalizes to languages unseen during alignment.
Furthermore, 
we show that separately trained alignment modules can be merged with existing task-specific modules, improving cross-lingual capabilities without full re-training.
Our code is publicly available\footnote{\url{https://github.com/dannigt/mid-align}}.
%
\end{abstract}

\section{Introduction}
Decoder-only large language models (LLMs) have emerged as the dominant paradigm in NLP.
While these models exhibit promising zero-shot capabilities \cite{DBLP:conf/iclr/WeiBZGYLDDL22,palm2}, 
further task-specific fine-tuning remains crucial for optimal performance in many applications \cite{DBLP:conf/icml/ShenTHAF24,DBLP:conf/iclr/Xu0SA24,DBLP:journals/corr/abs-2402-17733}.
During fine-tuning,
a practical challenge is that the available training data rarely covers all languages supported by LLMs.
This highlights the importance of cross-lingual transfer to extend task-specific performance gains across languages.

While cross-lingual transfer has been extensively studied \cite{DBLP:conf/apsipa/WangZ15,ruder-etal-2019-transfer,artetxe-schwenk-2019-massively,pfeiffer-etal-2020-mad},
achieving it on generative tasks with variable-length outputs
remains challenging \cite{vu-etal-2022-overcoming,li-murray-2023-zero} compared to classification tasks.
This challenge is especially relevant for LLMs, 
which formulate all tasks as next-token prediction problems.

The theoretical foundation of cross-lingual transfer lies in the analogous relationships between concepts across languages. 
This intuition was first demonstrated in cross-lingual word embeddings \cite{DBLP:journals/corr/MikolovLS13,DBLP:conf/iclr/LampleCRDJ18,xu2021crosslingualbertcontextualembedding}, 
where these vector representations exhibit isometric relationships, 
i.e., the geometric structure of semantically equivalent items is preserved across different languages.
This isometry property has proven crucial for transferring learned models across languages \cite{schuster-etal-2019-cross,wang2024bridginglanguagegapslarge}.
Subsequent encoder-decoder models \cite{ha2016multilingualneuralmachinetranslation} and decoder-only models \cite{wu-etal-2024-representational} also exhibit similar properties in their internal representations.

While pretrained multilingual models naturally develop some degree of unified multilingual representations  \cite{pires-etal-2019-multilingual, conneau-etal-2020-unsupervised,muller-etal-2021-first}, 
explicitly strengthening the relationships between semantically equivalent content has shown benefits in various downstream tasks:
cross-lingual 
retrieval \cite{yu-etal-2018-multilingual},
parallel text mining \cite{schwenk-etal-2021-wikimatrix}, 
zero-shot classification \cite{hu-etal-2021-explicit,gritta-iacobacci-2021-xeroalign} and translation \cite{arivazhagan2019missingingredientzeroshotneural,pham-etal-2019-improving,duquenne-etal-2022-modules}.
Despite different approaches,
these works share a common objective: 
\textit{aligning} representations of semantically equivalent content across languages while preserving overall expressiveness.

Cross-lingual alignment approaches have been successfully applied to models preceding LLMs.
For \textit{encoder-only} models, 
outputs can be aligned by e.g., minimizing distances between parallel sentence representations \cite{feng-etal-2022-language} or
cross-lingual masked language modeling objectives  \cite{DBLP:conf/nips/ConneauL19}.
These techniques are largely applicable to \textit{encoder-decoder} models, 
where alignment is typically enforced to the encoder outputs \cite{DBLP:journals/corr/abs-2308-11466}.
In contrast, \textit{decoder-only} models lack such clear separation between input processing and output generation.
This makes it less obvious where and how to optimize for cross-lingual alignment, 
as also highlighted in the survey by \citet{hammerl-etal-2024-understanding}.

In this work,
we start by quantifying the degree of cross-lingual alignment present in two prominent LLMs, 
Llama 3 \cite{grattafiori2024llama3herdmodels} and Qwen 2.5 \cite{qwen2025qwen25technicalreport}. 
We then apply these insights to improve cross-lingual transfer in task-specific fine-tuning.
By alternatively training on alignment and task-specific data, we aim to improve the cross-lingual generalization to languages without fine-tuning data.
We demonstrate transfer improvements across diverse tasks: slot filling, machine translation, and structured text generation.
Our main findings include:
\begin{itemize}[nolistsep,leftmargin=*]
    \item Applying alignment objectives to middle layers during LLM task-specific fine-tuning improves cross-lingual transfer (\S\ref{subsec:gains_on_transfer})
    and enhances alignment across all network depths (\S\ref{subsec:loss_placement}).
    \item The transfer improvements extend beyond those languages seen in alignment (\S\ref{subsec:gains_on_transfer}).
    \item Our approach is robust to the choice of languages used for alignment training (\S\ref{subsec:res_level}, \ref{subsec:domain_lang_generalization}).
    \item Task-specific and alignment modules trained separately can be combined post-hoc to improve transfer performance (\S\ref{subsec:merging}).
\end{itemize}

\begin{figure}[t]
    \centering
    \begin{subfigure}[b]{1.01\linewidth}
         \centering
         \includegraphics[width=\linewidth,clip,trim={0 0.6cm 0 0}]{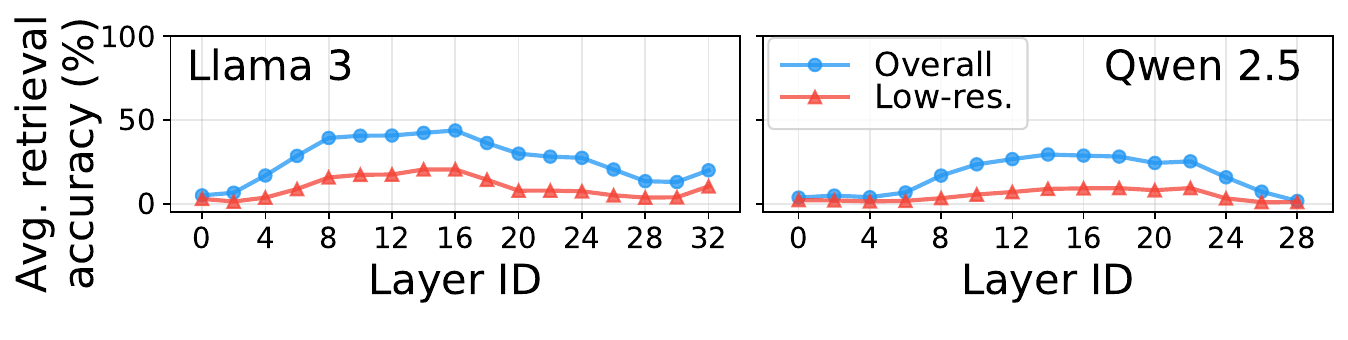}
         \caption{Cross-lingual semantic alignment (measured by average retrieval accuracy over 35 languages and 1190 language directions) varies by layer, with the middle layer showing the highest score. Lower-resource languages are poorly aligned.}
         \label{fig:retrieval_by_layer}
     \end{subfigure}
     
    \begin{subfigure}[b]{\linewidth}
         \centering
         \includegraphics[width=1.01\linewidth,clip,trim={0 0.1cm 0 0}]{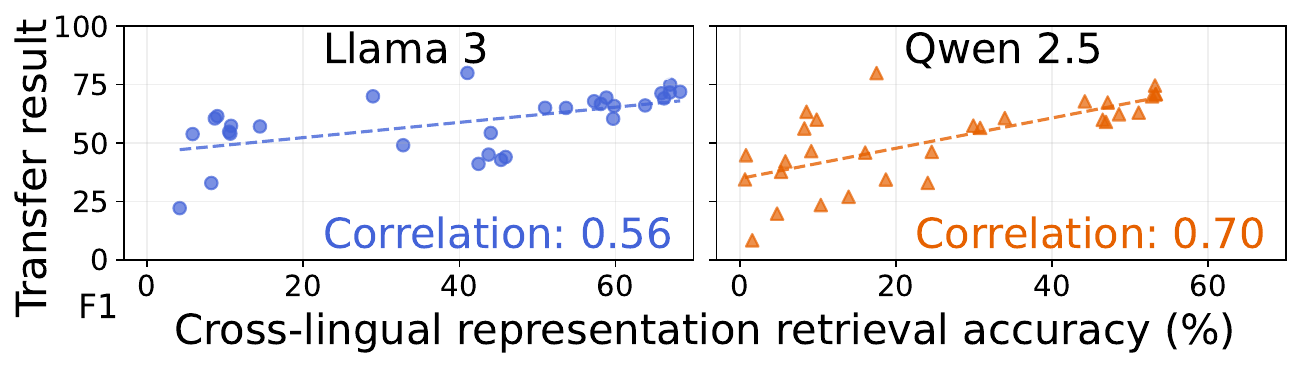}
         \caption{Positive correlation between base model cross-lingual semantic alignment and downstream transfer performance.}
         \label{fig:retrieval_vs_transfer_correlation}
     \end{subfigure}
    
    \caption{Two observations (\S\ref{sec:analysis}) motivating our approach of aligning multilingual representations (\S\ref{sec:approach}).}
    \label{fig:correlation}
\end{figure}

\section{Analyzing Cross-Lingual Alignment} \label{sec:analysis}

To understand how well LLM representations capture semantic equivalence across languages,
we use translation retrieval as a diagnostic task. 
We choose this retrieval task over other metrics like cosine similarity or SVCCA score \cite{DBLP:conf/nips/RaghuGYS17}
because it better captures \textit{relative} semantic relationships. 
That is,
if a model's representations enable us to identify a sentence's translation from a set of candidates,
the exact numerical distance between the query and the retrieved translation 
is less important than the ability to rank translations as the most semantically similar.

Specifically, 
we first extract model activations at each network layer for all language variants of the input text. 
To handle variable-length sequences, 
we create fixed-size sentence embeddings by mean-pooling the activations over the sequence length dimension. 
For translation retrieval, 
given a query sentence in one language, 
we compare its embedding to the embeddings of candidate sentences in the target language using ratio-based margin similarity \cite{artetxe-schwenk-2019-margin}\footnote{shown to outperform cosine similarity for cross-lingual retrieval tasks \cite{artetxe-schwenk-2019-margin}}. 
For $N$ languages, 
we evaluate retrieval accuracy across all $N(N-1)$ possible language pairs.
We use the \textsc{FLoRes-200} dataset \cite{DBLP:journals/nature/Team24}, 
which provides high-quality multiway parallel texts across diverse languages (detailed setup in \S\ref{subsec:setup_evaluation}).

Our investigation of LLama 3 and and Qwen 2.5 models\footnote{specifically the \texttt{8B-Instruct} and \texttt{7B-Instruct} variants} reveals three key findings:

\noindent
\textbf{Overall weak semantic alignment, with peak in middle layers:}
As shown in \autoref{fig:retrieval_by_layer}, 
the average translation retrieval accuracy across 1,190 language pairs remains below 50\%, 
with Llama~3 outperforming Qwen 2.5. 
Low-resource languages\footnote{resource levels as defined by \citet{DBLP:journals/nature/Team24}} show especially weak alignment, achieving less than half of the overall average accuracy. 
In particular, 
the \textit{middle} layers of both models demonstrate the strongest retrieval performance. 
This suggests stronger potential for cross-lingual transfer at these intermediate representations.

\noindent
\textbf{Strong correlation between base LLM semantic alignment and downstream task transfer:}
To what extent can the semantic alignment present in the base LLM predict cross-lingual transfer performance after supervised fine-tuning?
Using multilingual slot filling as a case study, 
we train models on 5 high-resource languages jointly and evaluate transfer performance on 25 additional languages 
(detailed setup in \S\ref{subsec:exp_setup_data}). 
As shown in \autoref{fig:retrieval_vs_transfer_correlation}, 
for both Llama 3 and Qwen 2.5, 
we observe strong positive correlations ($p<0.01$) between middle-layer retrieval accuracy and downstream task performance.
This correlation suggests that increasing cross-lingual alignment in LLM intermediate representations may improve cross-lingual transfer.

\begin{figure}[t]
    \centering
         \centering
         \includegraphics[width=\linewidth,clip,trim={0 0.6cm 0 0}]{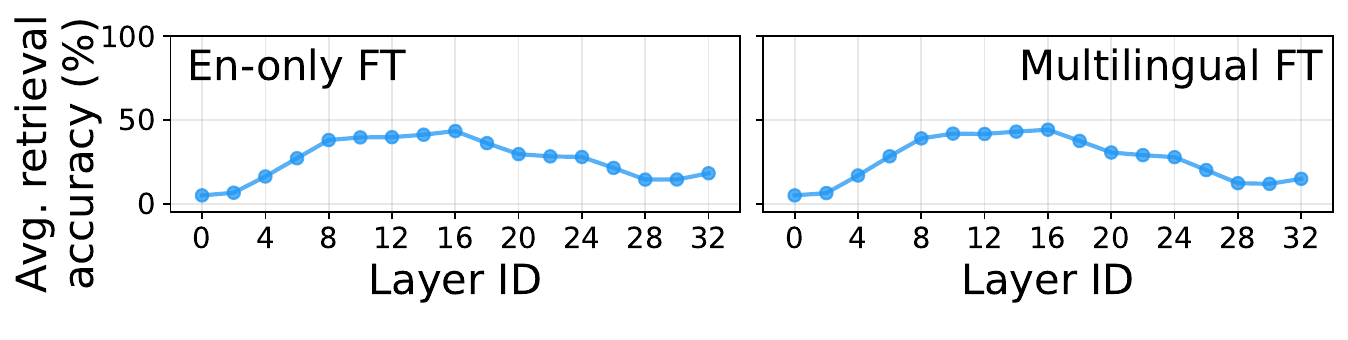}
         \vspace{-20pt}
         \caption{Task-specific fine-tuning shows minimal impact on semantic alignment.}
    \label{fig:FT_impact_on_retrieval}
\end{figure}

\noindent
\textbf{Task-specific fine-tuning preserves but does not enhance semantic alignment:}
After analyzing the base LLMs, 
we examine how supervised fine-tuning affects the models' internal semantic alignment.
Using the same multilingual slot filling task as before, 
we study both English-only and multilingual fine-tuning. 
Despite multilingual fine-tuning being an established method for improving cross-lingual transfer \cite{li-murray-2023-zero,chirkova-nikoulina-2024-key}, 
we observe that neither training configuration alters the models' cross-lingual semantic alignment (\autoref{fig:FT_impact_on_retrieval}). 
This preservation of baseline alignment patterns,
even under multilingual training,
indicates that pure fine-tuning does not sufficiently strengthen cross-lingual alignment. 
This further motivates us towards explicit cross-lingual alignment during fine-tuning.

\begin{figure*}[ht!]
    \centering
    \includegraphics[width=1\linewidth,clip,trim={4.5cm 3cm 2cm 0.65cm}]{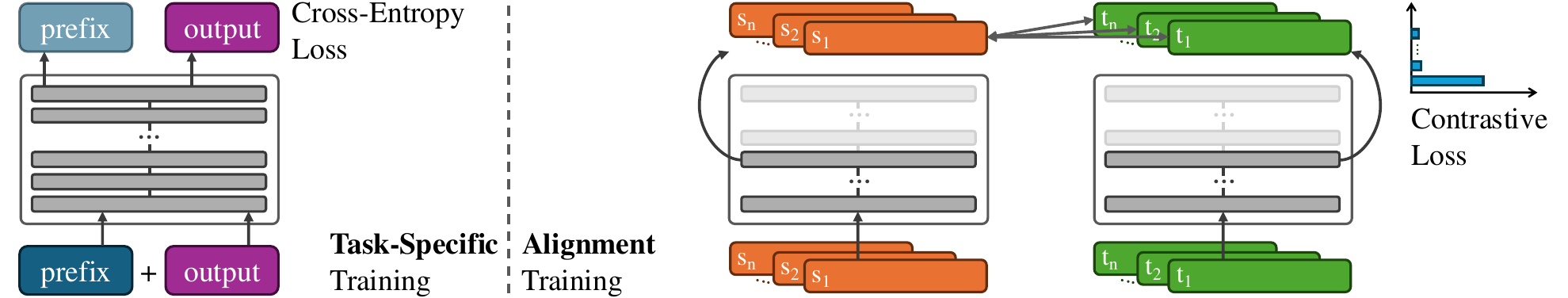}
    \caption{Illustration of our approach, alternating training between task-specific (left) and alignment (right) objectives. 
    The alignment objective operates on middle-layer representations.}
    \vspace{-10pt}
    \label{fig:overall_approach}
\end{figure*}

\section{Explicit Alignment in fine-tuning} \label{sec:approach}
\textbf{Alternate Training}
As shown in \autoref{fig:overall_approach},
we optimize either the task-specific objective or the alignment objective in each training step.
Compared to joint optimization that computes a combined loss for both objectives and performs a single backward pass, 
this approach does not involve manually balancing objective weights 
and mitigates potential gradient conflicts between objectives.
It also showed stronger task performance empirically.

\noindent
\textbf{Task Objective} 
We follow standard causal language modeling, 
using a cross-entropy loss over the predicted text conditioned on the input prefix.

\noindent
\textbf{Alignment Objective} 
We use a contrastive loss motivated by its successful applications in sentence embedding \cite{feng-etal-2022-language},
dense retrieval \cite{karpukhin-etal-2020-dense}
and modality alignment \cite{ye-etal-2022-cross,DBLP:conf/cvpr/GirdharELSAJM23}.
The loss maximizes the similarity between translations while minimizing similarity between non-translations.
Given a batch $\mathcal{B}$ of $n$ pairs of parallel sentences, the alignment loss for a sentence pair $(s,t)$ is:
\begin{equation}
\mathcal{L}_{\text{align}} = -\log \frac{\exp(\text{sim}(\mathbf{h}_s^i, \mathbf{h}_t^i))}{\sum_{v\in\mathcal{B}} \exp(\text{sim}(\mathbf{h}_s^i, \mathbf{h}_v^i))}
\end{equation}
where $\mathbf{h}_s^i$ is the mean-pooled\footnote{Initial experiments with attention pooling degraded performance.
We also tried a stop-gradient operator on English representations to  align non-English representations towards English, but it did not give consistent gains.} hidden states at the $i^{th}$ LLM layer for input $s$ and 
$\text{sim}(\cdot,\cdot)$ is a similarity function.
Motivated our finding that middle layers have the strongest cross-lingual alignment potential, 
we select $i$ as the middle layer
and compare its performance to other layer positions.
We use cosine similarity following prior works \cite{gao-etal-2021-simcse,ye-etal-2022-cross}.
The similarity score is optionally scaled by a temperature parameter $\tau$,
which controls the peakiness of the softmax distribution and in turn determines the relative importance of non-translation pairs.
This temperature parameter is tuned on the development sets.

\noindent
\textbf{Activating Individual Objectives} 
Note that the task and alignment losses can be activated separately. 
Deactivating the alignment loss degenerates to standard task-only training. 
Conversely, deactivating the task loss trains the model only for alignment.
The modularity allows combining separately-trained task and alignment models.

\noindent
\textbf{Data Requirement}
Our approach requires minimal parallel data.
Later experiments show that 
for lower-resource languages, 
a few hundreds of sentences of parallel data is sufficient to improve transfer.
Our approach also offers a practical advantage over alternatives that require monolingual language modeling training for each transfer target language \cite{ansell-etal-2022-composable,vu-etal-2022-overcoming,chronopoulou-etal-2024-language}.

\section{Experimental Setup} \label{sec:exp_setup}

\subsection{Data} \label{subsec:exp_setup_data}
\begin{table}[t]
\small
\centering
\setlength\tabcolsep{2pt} 
\begin{tabular}{llccccc}
\toprule
\multicolumn{2}{l}{} & 
\textbf{Dataset} & 
\textbf{Languages} & 
\\
\midrule
\multicolumn{4}{l}{\textbf{Slot Filling}} \\
& Task - train 
& \multirow{1}{*}{\shortstack[l]{MASSIVE}}
& \{ar, en, es, ru, zh\}
\\
\cmidrule{4-5}
& \multirow{2}{*}{\shortstack[l]{Task - test}}
& \multirow{2}{*}{\shortstack[l]{MASSIVE}}
& supervised + \{af, az, cy, de, el,  \\
&
&
&
fr, hi, is, ja, jv, sw, th, tl, tr, ur\}
\\
\cmidrule{4-5}
& \multirow{3}{*}{\shortstack[l]{Alignment}}
& \multirow{3}{*}{\shortstack[l]{Tatoeba}}
& low-res.: \{cy, jv, jp, sw, tl\}-en
\\
&
&
& mid-res.: \{el, hi, th, tr\}-en
\\
&
&
& high-res.: \{ar, es, ru, zh\}-en
\\
\midrule
\multicolumn{4}{l}{\textbf{Machine Translation}} \\
& Task - train 
& ALMA
&  \{cs, de, is, ru, zh\}$ \leftrightarrow$ en
\\
& Task - test
& WMT 23
& supervised + \{he, ja, uk\} $\leftrightarrow$ en
\\
& Alignment
& \multicolumn{2}{c}{{(same as ``Task - train'')}}
\\
\midrule
\multicolumn{4}{l}{\textbf{JSON Generation} (challenge task)}  \\
& Task - train 
& UNER
&  \{en, pt, zh\}
\\
& Task - test
& UNER
&  supervised + \{da, hr, sk, sr, sv\}
\\
& Alignment
& Tatoeba
&  \{da, sv\}-en
\\
\midrule
\multicolumn{4}{l}{\textbf{Semantic Alignment Evaluation} (diagnostic task)}  \\
&
Alignment &
FLoRes-200 &
$N(N-1)$ pairs for $N$ lang.
\\
\bottomrule
\end{tabular}
\caption{\label{tab:data_overview} 
Dataset statistics for three downstream tasks and one diagnostic task.
``Train'' refers to languages involved in SFT, and "test" includes SFT languages and additional transfer languages unseen during training.
See \autoref{sec:appendix_dataset_details} for more details.
}
\end{table}

In general, we fine-tune on several 
high-resource languages and then evaluate transfer performance on additional languages.
We do not focus on English-only fine-tuning, since our initial experiments demonstrated that multilingual fine-tuning substantially outperforms English-only fine-tuning\footnote{These English-only FT results are in Appendix~\ref{sec:appendix_english_only}.}, 
thus establishing it as a stronger baseline.
\autoref{tab:data_overview} presents a dataset overview.
Descriptions of the language codes are in \autoref{sec:appendix_language_list}.

\noindent
\textbf{Main Task Data:} 
We evaluate our approach on slot filling and machine translation, 
both modeled as generative tasks with templates shown in Appendix~\ref{sec:appendix_prompt_format}.
For slot filling, 
we use the \textsc{MASSIVE} dataset \cite{fitzgerald-etal-2023-massive}.
We train on 5 high-resource languages, 
and evaluate transfer performance on 15 additional diverse languages, 5 of which have non-Latin writing systems.
This task presents a challenge due to the 60 possible slots,
requiring strictly following the output format for correct parsing.
For machine translation, we use ALMA \cite{DBLP:conf/iclr/Xu0SA24}'s training and test data, and additionally test on 6 zero-shot directions from \textsc{WMT 23} \cite{kocmi-etal-2023-findings}.

\noindent
\textbf{Challenge Task Data:}
To assess performance on long-sequence processing and structured text generation,
we include JSON generation as a challenge task.
We use the UNER dataset \cite{mayhew-etal-2024-universal} from the Aya collection \cite{singh-etal-2024-aya}, 
which requires following example instructions and extracting named entities into JSON format.
A challenge not present in the previous tasks is the longer inputs, with an average input length exceeding 150 tokens in English. 
For this task,  we train on 3 high-resource languages (en, pt, zh) and transfer to the 5 remaining languages.

\noindent
\textbf{Alignment Data:}
For alignment, we mainly use parallel data to English from Tatoeba \cite{tiedemann-2020-tatoeba}, 
except for machine translation, where the training sentences are inherently parallel.
For slot filling, our main experiments align the five languages with the weakest baseline\footnote{their baseline is an XLM-R model trained on English} transfer performance (cy, jv, jp, sw, tl) reported by the dataset creators \cite{fitzgerald-etal-2023-massive}.
We choose them because their weak baseline performance suggests a lack of effective transfer, providing a strong testbed for evaluating the potential benefits of our alignment approach.
For ablation, we alter the following factors of the alignment data:
\begin{itemize}
    \item Resource level (low, medium, high-resource)
    \item Language coverage 
    \item Domain (oracle data, different, very distant)
\end{itemize}
For machine translation, given the inherent semantic equivalence of translation pairs, we directly leverage the translation data for alignment.
For JSON generation, we align the two lowest-resourced in UNER (da and sv)\footnote{While 
Serbian (sr) is also low-resourced in UNER, 
we exclude it from alignment due to data quality.
Running language identification reveals that many sentences in the Serbian alignment data are not actually in Serbian.} to English.
For lower-resource languages, the alignment data are a few hundreds as detailed in \autoref{sec:appendix_dataset_details}.

\subsection{Evaluation} \label{subsec:setup_evaluation}
\noindent
\textbf{Semantic Alignment Evaluation:}
As described in \S\ref{sec:analysis}, we evaluate cross-lingual semantic alignment by retrieval accuracy. 
Given $N$ languages, we perform many-to-many retrieval and average the accuracy over the $N(N-1)$ language pairs.
For the initial analyses (\S\ref{sec:analysis}), the 35 languages are listed in Appendix \ref{sec:appendix_language_list}.
We use the FLoRes-200 \cite{DBLP:journals/nature/Team24} development set with 997 parallel sentences. 
While FLoRes partially overlaps with ALMA's training data, 
it remains the only reliable massively multilingual multiway corpus to the best of our knowledge. 
Alternative such as Tatoeba have been advised against due to data imbalance and noise \cite{heffernan-etal-2022-bitext,janeiro2024mexmatokenlevelobjectivesimprove}. 
We also demonstrate that this overlap does not result in memorization effects (\S\ref{subsec:domain_lang_generalization}).
When reporting an aggregated retrieval accuracy for a model, 
we average over all language pairs at even-numbered layers' retrieval accuracy, 
excluding the input embedding layer.

\noindent
\textbf{Task Performance Evaluation:}
For slot filling, we report F$_1$ scores using the original evaluation script by \citet{fitzgerald-etal-2023-massive}.
For machine translation, we report BLEU\footnote{nrefs:1|case:mixed|eff:no|tok:13a|smooth:exp|version:2.4.2
sacreBLEU \cite{post-2018-call} signature, 
with "tok:ja-mecab-0.996-IPA" for Japanese and "tok:zh" for Chinese.} \cite{papineni-etal-2002-bleu} and COMET-22 \cite{rei-etal-2022-comet} scores.
For JSON generation, 
we parse the generated outputs back to named entity tuples and then evaluate F$_1$ scores.

\subsection{Model, Training, and Inference} 
We build upon Llama \cite{grattafiori2024llama3herdmodels} and Qwen \cite{qwen2025qwen25technicalreport}, specifically 
\texttt{Meta-Llama-3-8B-Instruct}\footnote{chosen over more recent versions to limit test set contamination, as its knowledge cutoff (March 2023) predates our translation test set (WMT 23).}
and \texttt{Qwen2.5-7B-Instruct}. 
We use LoRA \cite{DBLP:conf/iclr/HuSWALWWC22} adapters with a rank of 8 for all attention components and linear projections.
The effective batch size is 128 for both objectives, 
with mini-batches of 32 examples considered for the contrastive objective\footnote{While contrastive learning typically benefits from larger batch sizes  \cite{DBLP:conf/nips/ChenZXCD0TZC22}, our initial experiments with increased batch sizes did not give consistent improvements.}. 
Alignment data from different languages are re-sampled to an approximately uniform distribution. 
More details are in \autoref{sec:appendix_training_details}.

\section{Main Results} \label{sec:main_res}
The main results are summarized in \autoref{tab:overall_results}. 
Before assessing our proposed approach, 
we first establish the necessity of supervised FT by comparing it with zero-shot usage of the LLMs (rows $(2,5)$ vs. $(1,4)$).
On slot filling, the zero-shot performance of Llama 3 is very poor, 
achieving only 6.6\% F$_1$ on English due to difficulties in adhering to task-specific formats.
We therefore do not evaluate its zero-shot performance on all languages.
In machine translation, supervised fine-tuning shows substantial gains of 4-6 COMET over zero-shot.

\begin{table*}[ht!]
\small
\centering
\setlength\tabcolsep{1pt} 
\begin{tabular}{clcccccccccccccccccccccc}
\toprule
\textbf{ID} &
\textbf{Model} &
\multicolumn{4}{c}{
\textbf{Slot Filling} (\textsc{MASSIVE})
} 
&&
\multicolumn{7}{c}{
\textbf{Machine Translation} (\textsc{WMT}23)
} 
\\
\cmidrule{3-6} 
\cmidrule{8-14}
&& 
Supervised &
Transfer & 
Transfer &
Retrieval & 
& 
\multicolumn{2}{c}{
Supervised
} 
&
\multicolumn{4}{c}{
Transfer %
} 
& 
Retrieval 
\\
&&
\multicolumn{1}{c}{
(5 lang.)
}  &
\multicolumn{1}{c}{
(15 lang.)
}  &
(aligned) &
\multicolumn{1}{c}{
(all 20 lang.)
} 
&
& 
\multicolumn{2}{c}{
(5 lang.$\leftrightarrow$En)
}  &
\multicolumn{2}{c}{
(3 lang.$\rightarrow$En)  
}  &
\multicolumn{2}{c}{
(En$\rightarrow$3 lang.)  
}  &
\multicolumn{1}{c}{
(all 9 lang.) 
} 
\\
\cmidrule{3-6} 
\cmidrule{8-14}

&&
F$_1$ &
F$_1$ &
F$_1$ &
Acc. &
 &
BLEU &
COMET &
BLEU &
COMET &
BLEU &
COMET &
Acc.
\\
\midrule
$(1)$ & 
\textsc{Llama 3} &
-- & 
-- &
-- &
39.1 &
&
25.8 & 75.5 &
27.8 & 75.8 &
14.8 & 71.3 &
51.5
\\
$(2)$ & 
$+$ SFT &
76.6 & 
60.2 &
51.7 &
39.4 &
&
30.0 & 81.5 &
31.8 & 82.8 &
15.5 & 79.6 &
(55.3)
\\
$(3)$ & 
\phantom{00}$+$ alignment &
\textbf{77.0} &
\textbf{61.7} &
\textbf{55.5} &
73.2
&
&
29.9 & 81.5 &  
\textbf{32.3} & 83.0 &
\textbf{17.0} & \textbf{80.7}  & 
(84.5)
\\
$(4)$ & 
\textsc{Qwen 2.5} &
-- & 
-- &
-- &
21.4 &
&
23.0 & 74.5 &
28.5 & 81.3 &
12.6 & 71.2 &
36.5
\\
$(5)$ & 
$+$ SFT &
76.3 & 
53.5 &
41.6 &
20.9 &
&
27.4 & 78.4 &
29.7 & 82.7 &
14.6 & 76.9 &
(38.8)
\\
$(6)$ & 
\phantom{00}$+$ alignment &
\textbf{77.0} & 
\textbf{55.3} &
\textbf{46.5} &
{20.5} &
&
27.2 & 77.6 & 
\textbf{30.8} & 82.7 &
14.7 & 76.9 &
(75.6)
\\
\bottomrule
\end{tabular}

\caption{\label{tab:overall_results}
Overall supervised and transfer results.
Retrieval accuracy are averaged over all language pairs and layers.
\textbf{Bold}: highest task scores which outperforms the other setups.
(Results in brackets): potentially inflated scores due to partial overlap between retrieval and translation data.
Language-specific results are in \autoref{sec:appendix_individual_languages}.
}
\vspace{-10pt}
\end{table*} 
\subsection{Overall Performance Comparison} \label{subsec:gains_on_transfer}

\noindent
\textbf{Gains in cross-lingual transfer with supervised performance preserved:}
Our approach improves cross-lingual transfer across different tasks and models. 
For slot filling, 
we observe gains in both supervised and transfer (F$_1$ $+$0.4 and $+$1.5 respectively) settings on Llama fine-tuning, 
with similar improvements on Qwen (F$_1$ $+$0.7 supervised, $+$1.8 transfer). 
In machine translation with Llama in row $(3)$, 
our approach brings substantial gains when transferring to out-of-English directions ($+$1.5 BLEU, $+$1.1 COMET).\footnote{The observation of alignment not improving supervised directions is in line with \citet{pan-etal-2021-contrastive}, 
where the purely contrastive learning setup also does not improve supervised scores over their baseline ("m-Transformer").}
For into-English directions, 
there is a modest improvement in $+$0.5 BLEU and $+$0.2 COMET.
The larger gains on out-of-English directions suggest the approach is more beneficial for non-English generation in this case.
For Qwen in row $(6)$, 
our approach shows minor gains in into-English translation ($+$1.1 BLEU but no change in COMET), and does not influence out-of-English scores.
It also leads to a degradation ($-$0.8 COMET) on supervised directions.
This is potentially due to Qwen's non-English-centric pretraining combined with our English-centric alignment data.
With this exception, 
our approach maintains or improves supervised performance while enhancing transfer.

\noindent
\textbf{Aligned languages improve the most, but gains extend to other languages:}
The diverse language coverage in the slot filling dataset allows us to 
compare how the alignment objective benefits transfer to both aligned and non-aligned languages.
While aligned languages show the strongest improvements (F$_1$ $+$4.2 and $+$4.9 for Llama and Qwen respectively),
the benefits extend to other languages. 
Over the remaining 10 non-aligned languages, 
there is an average F$_1$ improvement of 0.4 (per-language results in Appendix~\ref{sec:appendix_individual_languages}).
This suggest that the alignment step enhances the model's general cross-lingual transfer capabilities rather than optimizing for specific language pairs. 

\noindent
\textbf{Smaller gains on non-Latin script languages:}
Beyond overall performance improvements,
we observe smaller gains on languages with diverse writing systems.
Specifically, 
for the non-Latin script transfer languages in the slot filling task (Greek, Hindi, Japanese, Thai, Urdu), 
the average improvement is only 0.5 F$_1$ in contrast to the overall average gain of 1.5.
This reduced gain is likely related to suboptimal tokenization for these languages in multilingual models \cite{rust-etal-2021-good,DBLP:conf/nips/PetrovMTB23,hong-etal-2024-accelerating}.
When tokens poorly align with linguistic units, 
the mean-pooled sentence representations may poorly capture semantics, 
thereby impacting our alignment objective. 

\subsection{Alignment Loss Placement} \label{subsec:loss_placement}
To validate our choice of middle-layer alignment motivated by the analysis in \S\ref{sec:analysis},
we compare performance when applying the alignment loss at different network depths: bottom (8\textsuperscript{th}), middle (16\textsuperscript{th}), and top (32\textsuperscript{nd}) layers of Llama. 

\noindent
\textbf{Middle layer placement achieves more balanced improvements in transfer languages:}
As shown in \autoref{tab:loss_position},
compared to the "middle" configuration, 
the "bottom" configuration clearly leads to poor overall performance in both supervised and transfer settings, 
with a particularly strong degradation on the slot filling task.
While top-layer alignment maintains overall strong performance, 
it shows more unbalanced gains across transfer languages, 
as evidenced by the higher standard deviation of performance changes on transfer languages. 

\begin{table}[t!]
    \small
    \centering
    \setlength\tabcolsep{2.2pt}
    \begin{tabular}{l c c c c c c ccc}
    \toprule
    &\textbf{Supervised{\tiny $\uparrow$}} 
    & \textbf{Transfer{\tiny $\uparrow$}}
    & \textbf{Transfer SD{\tiny $\downarrow$}} 
    \\
    \midrule
    \multicolumn{4}{l}{
    \textbf{Slot filling} (\textsc{MASSIVE}): F$_1$}  
    \\
    SFT baseline &
    76.6 &
    60.2 &
    $-$
    \\
    Middle (layer 16) &
    77.0 & 
    61.7 &
    2.6 
    \\
    Top (layer 32) &
    76.6 & 
    62.0 &
    3.3 
    \\
    Bottom (layer 8) &
    76.8 & 
    58.0 &
    2.9 
    \\
    \midrule
    \multicolumn{4}{l}{
    \textbf{Machine translation} (\textsc{WMT}23): COMET
    } 
    \\
    SFT baseline & 
    81.5 &
    79.6 &
    $-$
    \\
    Middle (layer 16) &
    81.5 & 
    80.7 &
    3.7 
    \\
    Top (layer 32) &
    82.0 & 
    80.2 &
    4.2
    \\
    Bottom (layer 8) &
    81.2 & 
    80.1 &
    5.6
    \\
    \bottomrule
    \end{tabular}
    \caption{Impact of alignment loss placement on Llama~3. 
    Last column: standard deviation of gains on transfer languages compared to baseline.
    ``Top'' leads to more uneven gains across languages,
    while ``bottom'' degrades both supervised and transfer performance.\label{tab:loss_position}}
\end{table}

\begin{figure}[t]
    \centering
    \includegraphics[width=\linewidth,clip,trim={0 0.2cm 0 0}]{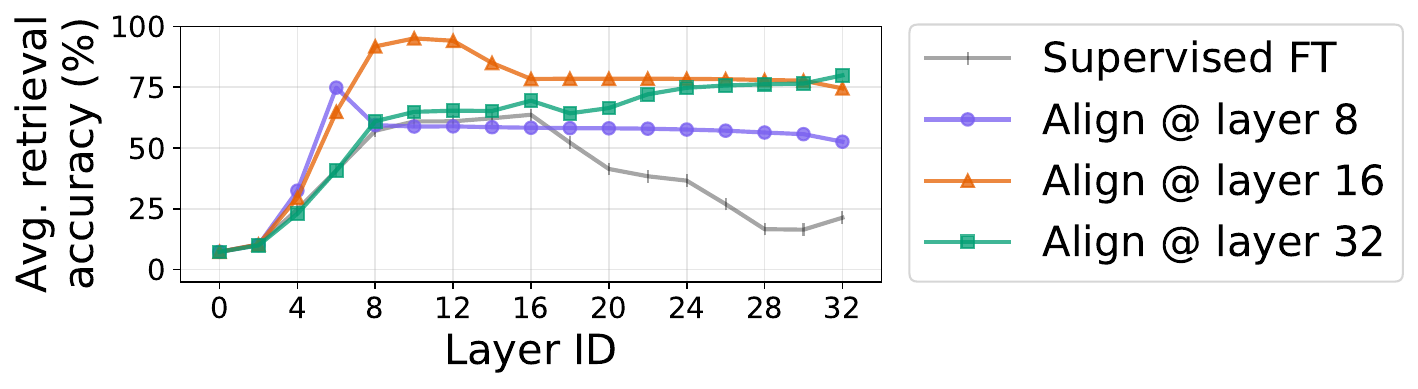}
    \caption{Retrieval accuracy over model depths when aligning different layers of Llama 3. Middle layer placement leads to overall better alignment.}
    \label{fig:ablation_loss_layer}
\end{figure}

\noindent
\textbf{Middle layer placement achieves better alignment across network depths:}
To better understand the effects of different loss placements, 
we run the translation retrieval task over model activations at from different intermediate layers.
As shown in \autoref{fig:ablation_loss_layer},
When the alignment loss is applied at the middle (16\textsuperscript{th}) layer, 
semantic alignment is enhanced not only at that layer but also in multiple preceding layers. 
In contrast, 
top-layer alignment primarily affects only the final layer, 
and bottom-layer alignment shows limited improvement in alignment quality across all layers. 
This is likely because the lower layers are occupied with processing more fundamental text features \cite{belinkov-etal-2017-neural,peters-etal-2018-deep} rather than abstract semantic meanings.

\noindent
\textbf{Aligning several layers does not show consistent gains:}
Results in \autoref{tab:loss_position} suggest that aligning at multiple layers may be complementary.
In Appendix \ref{sec:appendix_align_several_layers}, 
we show that adding alignment losses at both middle and top layers brings further improvements on slot filling, 
but does not on machine translation.
This task-dependent behavior indicates that how to best align multiple layers still requires further investigation.

\subsection{Impact on Representation Retrieval}
To assess the impact of the alignment loss on the learned model representations, 
we also report the retrieval accuracy for all languages involved in each task (20 for slot filling and 9 for machine translation) after fine-tuning in \autoref{tab:overall_results}.
For Llama on the slot filling task, the alignment loss substantially improves retrieval accuracy from 39.4\% to 73.2\%.
{For Qwen, the alignment loss does not improve retrieval among the 20 slot filling languages,
possibly due to the lower accuracy of the base model with many low-resource languages with 0\% accuracy,
making improvement more challenging.}
For machine translation, 
as noted earlier \S\ref{subsec:setup_evaluation}, the retrieval test data overlaps with part of the task training data, 
potentially inflating accuracy (marked in brackets in \autoref{tab:overall_results}).
However, we verify that this overlap does \textit{not} lead to perfect retrieval accuracy: 
Specifically, at the 16\textsuperscript{th} layer where the alignment loss is applied, 
English-source retrieval accuracies for supervised languages show varying accuracy:
cs (98.1\%), de (96.5\%), is (66.9\%), ru (90.6\%), and zh (94.8\%).
This suggests that the overlap does not make the retrieval diagnostic task trivial.

\section{Analyses}

\begin{table}[t!]
    \small
    \centering
    \setlength\tabcolsep{2.5pt}
    \begin{tabular}{l c c c}
    \toprule
    \textbf{Alignment Setup}
    &\textbf{Avg supervised} 
    & \textbf{Avg transfer}
    \\
    & 
    F1{\tiny $\uparrow$} (5 lang.) &
    F1{\tiny $\uparrow$} (15 lang.) 
    \\
    \midrule
    5 lang$\leftrightarrow$en (\autoref{tab:overall_results} row 3) &
    77.0 & 
    61.7 \phantom{($+$1.9)}
    \\
    All$\leftrightarrow$English (38 pairs) &
    77.5 & 
    63.6 ($+$1.9)
    \\
    All$\leftrightarrow$all (238 pairs) &
    77.7 & 
    63.6 ($+$1.9)
    \\
    \bottomrule
    \end{tabular}
    \caption{Effects of larger-scale alignment configurations on slot filling. Aligning all non-English languages to English (+1.9 F1) outperforms our base configuration, 
    while fully connecting all 20 languages offers no additional gain beyond English-only alignment.
    \label{tab:larger_scale_align}}
\end{table}
\subsection{Larger-Scale Alignment}
While our main configuration for slot filling (\autoref{tab:data_overview}) allows studying performance on languages not involved in alignment, 
we also explore larger-scale alignment scenarios as oracle setups where all languages have parallel data. 
We conduct additional experiments on slot filling with two expanded configurations:
\begin{itemize}[nolistsep,leftmargin=*]
    \item All 19 non-English languages aligned to English (38 directional pairs)
    \item All 20 languages aligned to each other (238 pairs with alignment data from all 380 possible pairs).
\end{itemize}
The results in \autoref{tab:larger_scale_align} show that expanding alignment to all languages further improves transfer performance (F1 $+$1.9) in an oracle setup where every transfer language has alignment data.
However, multiway alignment data does not further improve transfer, 
suggesting that aligning to English implicitly creates multiway alignment effects.

\subsection{Resource Level of Alignment Languages} \label{subsec:res_level}
\begin{table}[t!]
    \small
    \centering
    \setlength\tabcolsep{1.1pt}
    \begin{tabular}{l c c c c}
    \toprule
    \textbf{Alignment Language}
    &\textbf{Super.} 
    & \textbf{Transfer}
    & \textbf{Gain on Aligned}
    \\
    \textbf{Resource Level} & 
    (5 lang.) &
    (15 lang.) &
    (4/5 lang.)
    \\
    \midrule
    SFT (row $(2)$ \autoref{tab:overall_results}) &
    76.6 &
    60.2 &
    --
    \\
    Low (row $(3)$ \autoref{tab:overall_results}) &
    77.0 & 
    61.7 & 
    $+$3.8 
    \\
    Medium  &
    77.8 & 
    61.4 & 
    $+$1.1 
    \\
    High  &
    77.6 & 
    60.4 & 
    $+$0.7 
    \\
    \bottomrule
    \end{tabular}
    \caption{Effect of alignment language resource levels on slot filling F1{\tiny $\uparrow$}. 
    In three groups of alignment languages: 
    low (\{cy, jv, jp, sw, tl\}), 
    medium (\{el, hi, th, tr\}), 
    and high-resource (\{ar, es, ru, zh\}), 
    languages involved in alignment consistently show improvements, 
    with the strongest gains ($+$3.8 F1) in the low-resource group.
    \label{tab:resource_level}}
\end{table}

In our main experiments, we selected the 5 languages with the weakest performance from the \textsc{MASSIVE} baseline \cite{fitzgerald-etal-2023-massive} for alignment. 
We now vary the resource level of the alignment languages
using a medium-resource group with \{el, hi, th, tr\}$-$en
and a high-resource group with \{ar, es, ru, zh\}$-$en, 
which also have supervised task training data. 
As shown in \autoref{tab:resource_level}, 
all three configurations improve F$_1$ scores for the languages involved in alignment. 
However, the low-resource group exhibit the largest gains ($+$3.8 F$_1$), 
indicating that our approach is most beneficial to languages with weaker initial performance.
Moreover, 
overall transfer gains relative to the SFT baseline diminish when using high-resource languages for alignment, 
likely because these languages already have well-aligned representations and aligning them provides little benefit to lower-resource languages in the transfer set.
Overall, 
the results show that our approach is robust to the choice of alignment languages,
but selecting initially poorly aligned languages could provide broader benefits across different languages.

\subsection{Generalization of Learned Alignment}
\label{subsec:domain_lang_generalization}
\autoref{tab:generalization} examines the language and domain generalization of our alignment component.
To isolate the effects of task-specific joint training, 
we train the models using only the alignment loss,
following the same setup as our previous experiments but without optimizing on task-specific data.
We then evaluate retrieval accuracy as described in \S\ref{subsec:setup_evaluation}.

\noindent
\textbf{Language Generalization:}
While our main experiments align multiple language pairs, 
we now use single languages for alignment. 
As shown in \autoref{tab:generalization} (upper portion),
that single-language alignment training leads to diminished performance compared to multilingual training. 
Interestingly, 
we see comparable accuracy drops regardless of which individual language is used for alignment,
suggesting that the gains of multilingual alignment come from the diversity of the training data rather than characteristics of individual languages.

\noindent
\textbf{Domain Generalization:}
To isolate the effects of multilinguality,
we focus on alignment between a single language pair (English-German). 
In \autoref{tab:generalization} (lower portion),
we first establish an oracle setup using models trained on FLoRes data (Wikipedia domain, 
overlapping with retrieval data).
We then compare to two setups where the alignment data come from other domains: 
Tatoeba (short sentences for language learning; different) and IWSLT 2017 (public speaking transcriptions; very distant).
While we observe a decrease in retrieval accuracy compared to the oracle setup, 
the results suggest that, to enforce alignment into the model, it is not strictly necessary to source alignment data from the same domain as the task-specific data.

\begin{table}[t]
    \small
    \centering
    \setlength\tabcolsep{15pt}
    \begin{tabular}{l c c c c}
    \toprule
    \textbf{Alignment Data}
    & \textbf{Overall} (20 lang.)
    \\
    \midrule
    Multi \{ar,es,ru,zh,{sw}\}-en &
    80.2
    \\
    \phantom{00}Only de-en  &
    71.9
    \\
    \phantom{00}Only es-en &
    72.9
    \\
    \phantom{00}Only zh-en &
    72.7
    \\
    \midrule
    de-en FLoRes (oracle) & 
    77.7
    \\
    \phantom{00}Tatoeba (different) &
    71.9
    \\
    \phantom{00}IWSLT (very distant) &
    68.5
    \\
    \bottomrule
    \end{tabular}
    \caption{
Alignment generalization across languages and domains. \textit{Upper}:  Multilingual training improves overall alignment.
\textit{Lower}: Impacts of alignment transfer reasonably across domains, with performance drops when training data differs from test domain.
    \label{tab:generalization}}
\end{table}

\begin{table}[t!]
    \small
    \centering
    \setlength\tabcolsep{6.5pt}
    \begin{tabular}{l c c c c}
    \toprule
    \textbf{Setup}
    &\textbf{Supervised} 
    & \textbf{Transfer}
    \\
    \midrule
    \multicolumn{4}{l}{
    \textbf{Slot filling} (\textsc{MASSIVE}): F$_1${\tiny $\uparrow$}}  
    \\
    SFT (row $(2)$ \autoref{tab:overall_results}) &
    76.6 \phantom{($+$0.0)}&  
    60.2 \phantom{($+$0.0)}& 
    \\
    Joint (row $(3)$ \autoref{tab:overall_results}) &
    77.0 ($+$0.4) & 
    61.7 ($+$1.5)
    \\
    Merge &
    76.9 ($+$0.3) & 
    61.3 ($+$1.1) 
    \\
    \midrule
    \multicolumn{4}{l}{
    \textbf{Machine translation} (\textsc{WMT}23): COMET{\tiny $\uparrow$}} 
    \\
    SFT (row $(4)$ \autoref{tab:overall_results}) &
    81.5 \phantom{($+$0.0)}&  
    79.6 \phantom{($+$0.0)}& 
    \\
    Joint (row $(5)$ \autoref{tab:overall_results}) &
    81.5 ($+$0.0) & 
    80.7 ($+$1.1)
    \\
    Merge &
    82.0 ($+$0.5) & 
    80.2 ($+$0.6) 
    \\
    \bottomrule
    \end{tabular}
    \caption{Performance comparison of merged alignment and task modules versus joint training. Post-hoc merging of separately-trained LoRA adapters achieves comparable improvements to joint training.\label{tab:merging}} 
\end{table}

\subsection{Merging Alignment and Task Modules} \label{subsec:merging}
Our previous experiments focused on models jointly trained on both task and alignment objectives.
However, in practice, 
it may be necessary to enhance existing task-specific models with cross-lingual capabilities, 
where joint re-training is infeasible due to computational constraints or unavailability of the original task training data.
Inspired by recent advances in model merging
\cite{DBLP:conf/nips/MatenaR22,DBLP:conf/iclr/IlharcoRWSHF23}, 
we explore the feasibility of combining separately-trained task and alignment modules.
We merge two sets of trained LoRA adapters by averaging their weights\footnote{We use a weighted average tuned on the development set (details in Appendix \ref{appendix:inference_details})}: 
the alignment module trained in isolation (\S\ref{subsec:domain_lang_generalization}), 
and task-specific modules (rows (2) and (5) in \autoref{tab:overall_results}).

\autoref{tab:merging} shows that this post-hoc merging brings comparable improvements comparable to joint training.
Moreover, the improvements are more evenly distributed across languages compared to the larger gains observed on languages used directly in alignment.
These results demonstrate that our alignment approach is modular and can be combined with existing task-specific models.

\begin{table}[t!]
    \small
    \centering
    \setlength\tabcolsep{8pt}
    \begin{tabular}{l c c c c}
    \toprule
    &\textbf{Supervised} 
    & \textbf{Transfer}
    & \textbf{Transfer}
    \\
    & 
    (en, pt, zh) &
    (da, sv) &
    (5 lang.)
    \\
    \midrule
    Llama SFT  &
    83.4 & 
    82.1 & 
    79.3
    \\
    $+$ alignment  &
    82.4 & 
    83.1 & 
    79.8
    \\
    \bottomrule
    \end{tabular}
    \caption{Results on JSON generation evaluated with F$_1$, showing modest gains for aligned languages but decreased performance for supervised languages. \label{tab:uner}}
\end{table}

\subsection{Long Sequence Processing}
\label{subsec:uner}
We investigate a more challenge task requiring longer input and output generation using UNER (\S\ref{subsec:exp_setup_data}).
As shown in \autoref{tab:uner}, 
while aligned languages still show improvements, 
the gains are more modest compared to previous experiments, 
with an F$_1$ increase of 1.0 on aligned languages and 0.5 across all transfer languages. 
Moreover, 
there is an average degradation of 1.0 F$_1$ on supervised languages, 
mainly due to the decline in Chinese ($-$2.2 F$_1$).  
We suspect that this is due to our sentence-level alignment objective operates on fixed-length representations, which creates conflicts with processing longer sequences. 
As Chinese is the only character-based language in the JSON generation dataset, which has roughly twice the number of tokens compared to English of equivalent content, the conflict could be more influential for Chinese.

\section{Related Works}
\paragraph{Multilingual Capabilities of LLMs}
LLM performance varies across languages due to imbalanced pre-training data volume.  
However, even predominantly English-centric models \cite{DBLP:journals/corr/abs-2307-09288} exhibit some degree of multilingual capability \cite{aycock-bawden-2024-topic,yuan-etal-2024-vocabulary},
potentially due to the unintentional ingestion of multilingual data during pretraining \cite{briakou-etal-2023-searching}.
Meanwhile, many recent LLMs have expanded their language coverage \cite{grattafiori2024llama3herdmodels,qwen2025qwen25technicalreport}.
Despite these inherent multilingual capabilities, 
extending them to downstream tasks in low-resource settings \cite{adelani-etal-2024-comparing,iyer-etal-2024-exploring} remains challenging. 

\paragraph{Multilingual Representation Alignment}
Enhancing meaningful cross-lingual relationships between model representations has been a well-studied area in the context of many tasks,
including intermediate tasks such as 
bilingual lexicon induction \cite{zhang-etal-2017-adversarial} and
sentence embeddings \cite{feng-etal-2022-language,li-etal-2023-dual},
as well as more direct applications like 
information retrieval \cite{DBLP:journals/tmlr/IzacardCHRBJG22} and translation \cite{pham-etal-2019-improving}.

Multilingual representation alignment can be achieved by various mechanisms,
such as similarity losses that push translations toward each other \cite{pham-etal-2019-improving}, 
contrastive losses \cite{DBLP:conf/cvpr/HadsellCL06} that additionally incorporate non-translation pairs, 
and adversarial losses \cite{DBLP:conf/icml/GaninL15} that remove language-specific signals. 
The cross-lingual transfer capabilities of these approaches is extensively documented in the literature. 
In particular, contrastive learning methods have shown promising results \cite{pan-etal-2021-contrastive, chi-etal-2021-infoxlm, qi-etal-2022-enhancing}.
Our contribution is not applying contrastive learning itself, but rather investigating how to effectively align multilingual spaces specifically in decoder-only models.

In the context of LLMs, 
\citet{wang2024bridginglanguagegapslarge} 
use linear projections learned offline to align non-English representations with English ones during decoding.
Our work differs in that our alignment objective is parameterized by the same weights as task-specific fine-tuning, 
and is directly applicable to multilingual fine-tuning.
\citet{wu-etal-2024-representational} align LLM top-layer representations specifically for the task of semantic textual similarity (STS).
Different from this work, they do not consider cross-lingual transfer in downstream tasks or explore intermediate LLM layers for alignment.

\paragraph{LLM Representation Analysis}
Several recent works have analyzed LLM internal representations with
geometric analysis of representation spaces \cite{razzhigaev-etal-2024-shape,lee2024multimodalfoundationmodelsencode},
probing classifiers \cite{wang-etal-2024-probing-emergence,li2025exploringmultilingualprobinglarge},
or logit lens analysis \cite{wu2024semantichubhypothesislanguage}.
Multiple studies \cite{wu2024semantichubhypothesislanguage, mao-yu-2024-tuning, zhong2024englishcentricllmslanguagemultilingual} reported higher representational similarity in middle layers in various evaluation settings, complementing our findings.
\citet{wu2024semantichubhypothesislanguage} identify “semantic hubs” in LLM middle layers that integrate information from various data types, 
while we focus specifically on cross-lingual representations rather than multi-modality.
\citet{mao-yu-2024-tuning} show that SFT on machine translation increases similarity between parallel sentences from the same MT corpus, 
while we show that SFT on a non-translation task does \textit{not} increase representation similarity, thereby motivating explicit alignment during SFT.
\citet{zhong2024englishcentricllmslanguagemultilingual}
measure pairwise similarity to English representations ("latent language") on high resource languages, 
while we focus on pairwise similarity across a more diverse set of language pairs.
\citet{kargaran2024mexamultilingualevaluationenglishcentric} use similarity between parallel sentences to estimate cross-lingual transfer capabilities.
Our analysis in \S\ref{sec:analysis} shares the same motivation, and we additionally show that actively enforcing alignment can improve transfer performance.

\section{Conclusion}
We presented a simple yet effective approach for enhancing cross-lingual transfer in LLMs through middle-layer representation alignment during fine-tuning. 
Our experimental results lead to several practical recommendations:
1) Aligning a few weakly-performing languages yields broad transfer benefits. 
A few hundreds of parallel sentences as alignment data are sufficient.
2) Alignment data can be sourced from different domains as the task.
3) Existing task-specific models can be enhanced with our approach via parameter merging
without the need of full re-training.

\newpage
\section*{Limitations}
\paragraph{Performance on languages with diverse scripts:}
As discussed in \S\ref{subsec:gains_on_transfer}, 
our approach shows smaller gains on languages non-Latin scripts. 
This limitation is likely related to fundamental tokenization challenges, where suboptimal token segmentation negatively impacts the quality of mean-pooled representations.
While our initial experiments on attention pooling did not lead to improvements, 
exploring more sophisticated pooling mechanisms could potentially address this challenge in future work.

\paragraph{Computational overhead during training:}
The alternating optimization between task and alignment objectives doubles the computational cost during training compared to standard fine-tuning. 
In computationally constrained settings, our merging approach (\S\ref{subsec:merging}), 
which separates task-specific and alignment training, 
should be prioritized.  
Given that alignment can be effectively performed using only a small number of parallel sentences (a few hundred per language), 
this modular approach can significantly reduce the overall computational cost.

\paragraph{Trade-offs between supervised and transfer performance in challenging scenarios:}
While our approach generally maintains or improves supervised task performance while improving transfer,
we observe degradation in supervised performance in two specific scenarios.
First, in structured text generation (\S\ref{subsec:uner}), the method shows reduced effectiveness and can impair supervised performance ($-$1.0 F$_1$), suggesting that our sentence-level alignment may interfere with the processing of longer, structured sequences.
Second, when applying the method to models with weak initial cross-lingual alignment (\S\ref{subsec:gains_on_transfer}), there could be a trade-off between improved transfer and supervised performance.

\section*{Acknowledgments}
We thank the reviewers for their feedback, as well as Felix Stahlberg and Google Research.
Part of this work was funded by the KiKIT (The Pilot Program for Core-Informatics at the KIT) of the Helmholtz Association.
The authors gratefully acknowledge the computing time provided on the high-performance computer HoreKa by the National High-Performance Computing Center at KIT (NHR@KIT). This center is jointly supported by the Federal Ministry of Education and Research and the Ministry of Science, Research and the Arts of Baden-Württemberg, as part of the National High-Performance Computing (NHR) joint funding program. HoreKa is partly funded by the German Research Foundation (DFG).

\bibliography{custom,anthology}

\appendix

\section{English-Only Fine-Tuning Results}
\label{sec:appendix_english_only}
\begin{table*}[ht]
    \small
    \centering
    \setlength\tabcolsep{1.7pt}
    \begin{tabular}{l c c c c c  c c c c c  ccccccccccccccccccccccccccc}
    \toprule
    & 
    ar &
    en &
    es &
    ru &
    zh &
    cy &
    ja &
    jv &
    sw &
    tl &
    af &
    az &
    de &
    el &
    fr &
    hi &
    is &
    th &
    tr &
    ur 
    \\
    \midrule
    English-only 
    &
59.8 &
82.5 &
82.4 &
65.8 &
61.6 &
60.3 &
39.7 &
37.8 &
39.8 &
57.5 &
60.3 &
39.6 &
71.1 &
64.8 &
68.2 &
62.1 &
39.2 &
75.3 &
52.9 &
49.9 &
    \\
    Multilingual 
    &
    75.5 &
81.7 &
74.5 &
77.6 &
73.8 &
44.0 &
65.8 &
41.0 &
42.8 &
65.0 &
66.0 &
49.0 &
75.0 &
69.4 &
71.9 &
70.0 &
45.0 &
79.9 &
60.4 &
57.1
    \\
    \bottomrule
    \end{tabular}
    \caption{Per-languages F$_1$ results on slot filling of English-only finetuning compared to multilingual fine-tuning on \{ar, en, es, ru, zh\}. 
    Multilingual fine-tuning shows stronger transfer performance. \label{tab:full_results_enonly_vs_multilingual}}
\end{table*}

\begin{table*}[h!]
    \small
    \centering
    \begin{tabular}{c c c c cc c}
    \toprule
    \textbf{Code} 
    & \textbf{FLoRes Code} 
    & \textbf{Full Name}
    & \textbf{Slot Filling}
    & \textbf{Machine Translation}
    & \textbf{JSON Generation}
    \\
    \midrule
    af &
    afr\_Latn &
    Afrikaans  &
    $\checkmark$ 
    \\
    \rowcolor{Gray}
    az &
    azj\_Latn &
    North Azerbaijani &
    $\checkmark$ &&
    \\
    ar &
    arb\_Arab &
    Modern Standard Arabic &
    $\checkmark$ 
    \\\rowcolor{Gray}
    cs &
    ces\_Latn &
    Czech &&
    $\checkmark$ &
    \\
    cy &
    cym\_Latn &
    Welsh&
    $\checkmark$ 
    \\
    \rowcolor{Gray}
    da &
    dan\_Latn &
    Danish &&&    $\checkmark$ 
    \\
    de &
    deu\_Latn & 
    German &
    $\checkmark$ &
    $\checkmark$ 
    \\
    \rowcolor{Gray}
    el &
    ell\_Grek &
    Greek &
    $\checkmark$ &&
    \\
    en &
    eng\_Latn &
    English &
    $\checkmark$ &
    $\checkmark$ &
    $\checkmark$ 
    \\
    \rowcolor{Gray}
    es &
    spa\_Latn &
    Spanish &
    $\checkmark$ &&
    \\
    fr &
    fra\_Latn &
    French &
    $\checkmark$ 
    \\
    \rowcolor{Gray}
    he &
    heb\_Hebr &
    Hebrew &&
    $\checkmark$ &
    \\
    hi &
    hin\_Deva &
    Hindi &
    $\checkmark$ 
    \\
    \rowcolor{Gray}
    hr &
    hrv\_Latn &
    Croatian &&&
    $\checkmark$ 
    \\
    is &
    isl\_Latn & 
    Icelandic &
    $\checkmark$ &
    $\checkmark$ 
    \\
    \rowcolor{Gray}
    ja &
    jpn\_Jpan&
    Japanese &
    $\checkmark$ &
    $\checkmark$ &
    \\
    jv &
    jav\_Latn &
    Javanese &
    $\checkmark$ 
    \\
    \rowcolor{Gray}
    pt &
    por\_Latn &
    Portuguese &&&
    $\checkmark$ 
    \\
    ru &
    rus\_Cyrl &
    Russian &
    $\checkmark$ &
    $\checkmark$ 
    \\
    \rowcolor{Gray}
    sk &
    slk\_Latn &
    Slovak &&&
    $\checkmark$ 
    \\
    sr &
    srp\_Cyrl &
    Serbian &&&
    $\checkmark$ 
    \\
    \rowcolor{Gray}
    sv &
    swe\_Latn &
    Swedish &&&
    $\checkmark$ 
    \\
    sw &
    swh\_Latn &
    Swahili &
    $\checkmark$ 
    \\
    \rowcolor{Gray}
    th &
    tha\_Thai &
    Thai &
    $\checkmark$ &&
    \\
    tl &
    tgl\_Latn &
    Tagalog &
    $\checkmark$ 
    \\
    \rowcolor{Gray}
    tr &
    tur\_Latn &
    Turkish &
    $\checkmark$ &&
    \\
    uk &
    ukr\_Cyrl &
    Ukrainian &&
    $\checkmark$ 
    \\
    \rowcolor{Gray}
    ur &
    urd\_Arab &
    Urdu &
    $\checkmark$ &&
    \\
    zh &
    zho\_Hans &
    Chinese (Simplified) &
    $\checkmark$ &
    $\checkmark$ &
    $\checkmark$ 
    \\
    \bottomrule
    \end{tabular}
    \caption{List of languages evaluated on different downstream tasks. \label{tab:full_language_list}}
\end{table*}

\autoref{tab:full_results_enonly_vs_multilingual} compares English-only and multilingual fine-tuning on \textsc{MASSIVE}.
Multilingual fine-tuning substantially outperforms English-only in cross-lingual transfer performance.

\section{Dataset Details}
\label{sec:appendix_dataset_details}
All our task training data are retrieved from HuggingFace\footnote{
MASSIVE: \url{https://huggingface.co/datasets/AmazonScience/massive}\\
ALMA: \url{https://huggingface.co/datasets/haoranxu/ALMA-Human-Parallel}\\
UNER: \url{https://huggingface.co/datasets/CohereForAI/aya_collection/viewer/templated_uner_llm}}.
The translation test sets are hosted by WMT\footnote{ \url{https://github.com/wmt-conference/wmt23-news-systems/tree/master/txt}}.
The alignment data are sourced from Tatoeba\footnote{\url{https://huggingface.co/datasets/Helsinki-NLP/tatoeba}}
with its default version of
\texttt{v2021-07-22} at the time of writing.
We filter out translations that are empty or include multiple sentences.
The lowest-resource alignment languages have a few hundred parallel sentences: 
Javanese (264), Swahili (371), Welsh (823).
The ablation de-en alignment data is from IWSLT 2017\footnote{\url{https://huggingface.co/datasets/IWSLT/iwslt2017}} \cite{cettolo-etal-2017-overview}.

\section{List of Languages} \label{sec:appendix_language_list}
The languages involved in our downstream tasks are listed in \autoref{tab:full_language_list}.
The 35 languages in the initial analyses in \S\ref{sec:analysis}
include all languages in slot fill and machine translation.
They additionally include the following languages: 
am (Amharic), bn (Bengali), it (Italian), hu (Hungrian), hy (Armenian),
id (Indonesian), kn (Kannada), ka (Georgian ), mn (Mongolian), km (Khmer), 
ko (Korean), and lv (Latvian).

\section{Training and Inference Details}
\label{sec:appendix_training_details}

\subsection{Training Hyperparameters}
Fine-tuning is performed using LoRA \cite{DBLP:conf/iclr/HuSWALWWC22} adapters with a rank of 8 for all attention components and linear projections (query, key, value, output, gate, up, down).
We set LoRA's $\alpha$ parameter to 16 and dropout to 0.1.
The number of trainable parameter is 20,971,520 on Llama 3, and 20,185,088 on Qwen 2.5.
We train at most 5 epochs on the task data. 
Training on all our tasks converged before reaching the max number of epochs.
The learning rate is set to 5e-4 with inverse square root schedule and warmup up ratio 0.03.
We save checkpoints and evaluate every 200 optimization steps,
and early stop if the development loss does not improve for 5 consecutive evaluations.
For the temperature parameter $\tau$ in the contrastive loss, 
we searched among \{0.1, 1.0, 1.5, 2.0\} based on development loss on machine translation.
For Llama we 0.1, for Qwen we use 1.5.

\subsection{Prompt Format}
\label{sec:appendix_prompt_format}
\paragraph{Slot Filling}
The system prompt is shortened from \citet{DBLP:conf/interspeech/0001G23}.\footnote{We stayed consistent to the original prompt text, preserving the typographical errors too.}
\begin{itemize}[nolistsep,leftmargin=*]
\item \textbf{System}: Given a command from the user, a voice assistant will extract entities essential for carry out the command. 
Your task is to extract the entities as words from the command if they fall under a predefined list of entity types.
\item \textbf{User}: wake me up at five am this week
\item \textbf{Assistant}: time: five am; date: this week
\item \textbf{User} (de): wecke mich in dieser woche um fünf uhr auf
\item \textbf{Assistant} (de): date: dieser woche; time: fünf uhr
\end{itemize}

For \textbf{zero-shot slot filling} experiments, we need to specify more requirements in the system prompt with the template also following \citet{DBLP:conf/interspeech/0001G23}:

{
Given a command from the user, a voice assistant like Siri or Olly will extract entities from the command that are essential for carry out the the command. For example, for a command about playing a specific song, the name of the song mentioned by the user would be an entity, falling under the type of “song name”.

Your task is to extract the entities as words from the command if they fall under any of the types given below according to the following description:

transport\_descriptor
house\_place
music\_album
sport\_type
playlist\_name
movie\_name
song\_name
place\_name
radio\_name
cooking\_type
weather\_descriptor
person
email\_folder
business\_type
audiobook\_author
transport\_type
general\_frequency
meal\_type
game\_name
device\_type
transport\_name
time\_zone
joke\_type
drink\_type
email\_address
food\_type
date
relation
currency\_name
ingredient
player\_setting
movie\_type
definition\_word
game\_type
list\_name
artist\_name
personal\_info
audiobook\_name
timeofday
transport\_agency
media\_type
podcast\_name
coffee\_type
business\_name
news\_topic
app\_name
podcast\_descriptor
color\_type
music\_genre
event\_name
time
change\_amount
alarm\_type
order\_type
music\_descriptor

Please give answers like:

1. person: john; contact\_field: phone number

2. transport\_app: uber; time\_of\_day: tonight; time: ten pm

3. None

4. music\_genre: jazz

etc., each taking a single line. The entity type must be one of the types given above, and the entity must be copied verbatim from the command. There could be zero, one, or multiple entities in a command.
}

\begin{table*}[t!]
    \small
    \centering
    \setlength\tabcolsep{2pt}
    \begin{tabular}{l c c c c c | c c c c c | ccccccccccccccccccccccccccc}
    \toprule
    & \multicolumn{5}{c}{\textbf{Supervised}}
    & \multicolumn{5}{c}{\textbf{Transfer (aligned)}}
    & \multicolumn{10}{c}{\textbf{Transfer (other)}}
    \\
    & 
    ar &
    en &
    es &
    ru &
    zh &
    cy &
    ja &
    jv &
    sw &
    tl &
    af &
    az &
    de &
    el &
    fr &
    hi &
    is &
    th &
    tr &
    ur 
    \\
    \midrule
    Llama  3 SFT 
    &
75.5 &
81.7 &
74.5 &
77.6 &
73.8 &
44.0 &
65.8 &
41.0 &
42.8 &
65.0 &
66.0 &
49.0 &
75.0 &
69.4 &
71.9 &
70.0 &
45.0 &
79.9 &
60.4 &
57.1
    \\
    $+$ align (middle)
    &
75.1 &
82.0 &
74.9 &
78.0 &
74.9 &
49.4 &
66.5 &
48.2 &
47.7 &
65.5 &
66.2 &
47.9 &
74.7 &
72.4 &
72.1 &
69.6 &
48.0 &
79.1 &
62.2 &
56.1 &
    \\
    Qwen 2.5 SFT &
74.7 &
81.1 &
74.0 &
77.5 &
74.1 &
27.0 &
67.3 &
32.9 &
23.5 &
57.4 &
58.9 &
45.9 &
74.6 &
63.3 &
70.8 &
60.0 &
34.4 &
79.9 &
59.9 &
46.5 &
    \\
    $+$ align (middle) &
    74.9 &
82.5 &
74.8 &
78.0 &
75.1 &
36.5 &
68.3 &
39.6 &
30.4 &
57.8 &
63.1 &
42.5 &
74.6 &
63.3 &
70.9 &
61.3 &
35.8 &
80.2 &
58.1 &
47.2 &
    \\
    \bottomrule
    \end{tabular}
    \caption{Per-languages F$_1$ results on slot filling. \label{tab:full_results_slot_filling}}
\end{table*}

\begin{table*}[h!]
    \small
    \centering
    \setlength\tabcolsep{2pt}
    \begin{tabular}{l c c c c c | c c c c c | ccc | cccccccccccccccccccccccc}
    \toprule
    & \multicolumn{5}{c}{\textbf{Supervised X$\rightarrow$En}}
    & \multicolumn{5}{c}{\textbf{Supervised En$\rightarrow$X}}
    & \multicolumn{3}{c}{\textbf{Transfer X$\rightarrow$En}}
    & \multicolumn{3}{c}{\textbf{Transfer En$\rightarrow$X}}
    \\
    & 
    cs &
    de &
    is &
    ru &
    zh &
    cs &
    de &
    is &
    ru &
    zh &
    he &
    ja &
    uk &
    he &
    ja &
    uk 
    \\
    \midrule
    &
    \multicolumn{16}{c}{BLEU} \\
    Llama  3 SFT 
    &
37.8 &
43.0 &
28.3 &
32.0 &
22.5 &
25.9 &
35.5 &
10.6 &
25.2 &
38.9 &
39.3 &
17.5 &
38.7 &
14.5 &
14.2 &
17.7 
    \\
    $+$ align (middle) &
38.4 &
43.1 &
29.1 &
32.4 &
23.0 &
24.7 &
34.7 &
10.9 &
24.4 &
38.1 &
39.8 &
18.8 &
38.4 &
16.0 &
15.6 &
19.5 
    \\
    Qwen 2.5 SFT &
36.1 &
40.8 &
20.5 &
30.6 &
23.2 &
21.5 &
33.7 &
6.8 &
25.3 &
45.3 &
34.6 &
18.9 &
35.6 &
13.3 &
17.6 &
13.0 
    \\
$+$ align (middle) &
36.6 &
41.4 &
21.2 &
30.9 &
24.0 &
20.5 &
32.7 &
4.8 &
25.0 &
45.3 &
36.3 &
19.4 &
36.8 &
12.7 &
17.8 &
13.5 
    \\
&
\multicolumn{16}{c}{COMET} \\
Llama 3 SFT &
85.2 &
84.9 &
81.0 &
82.4 &
79.7 &
84.3 &
81.8 &
68.7 &
83.3 &
84.2 &
83.6 &
79.8 &
85.1 &
75.7 &
83.5 &
79.7 
\\
 $+$ align (middle) &
 85.5 &
84.9 &
81.1 &
82.4 &
79.8 &
83.8 &
81.6 &
69.0 &
83.3 &
84.0 &
83.6 &
80.1 &
85.2 &
77.1 &
84.2 &
80.8 
 \\
Qwen 2.5 SFT &
84.8 &
84.7 &
74.1 &
82.6 &
80.2 &
80.8 &
80.6 &
52.0 &
83.3 &
86.1 &
82.3 &
81.3 &
84.5 &
70.7 &
85.5 &
74.6 
\\
$+$ align (middle) &
85.1 &
84.7 &
74.4 &
82.6 &
80.4 &
79.5 &
80.1 &
46.5 &
83.1 &
85.8 &
82.2 &
81.4 &
84.6 &
70.7 &
85.7 &
74.4 
\\
    \bottomrule
    \end{tabular}
    \caption{Per-languages BLEU and COMET results on machine translation. \label{tab:full_results_machine_translation}}
\end{table*}

\paragraph{Machine Translation}
\begin{itemize}[nolistsep,leftmargin=*]
    \item \textbf{System}: Translate the following sentences from English to German.
    \item \textbf{User}: Police arrest 15 after violent protest outside UK refugee hotel.
    \item \textbf{Assistant}: Polizei verhaftet 15 Menschen nach gewalttätigen Protesten vor einer Flüchtlingsunterkunft in Großbritannien
\end{itemize}

\paragraph{JSON Generation} 
\begin{itemize}[nolistsep,leftmargin=*]
    \item \textbf{User}: Please identify all the named entities mentioned in the input sentence provided below. Use only the categories: PER - person, ORG - organization, and LOC - location. Remember, nationalities are neither locations nor organizations, and organizations can represent other groups of people. Pay attention to the provided example. You should only output the results in JSON format, following a similar structure to the example result provided. Example sentence and results: Where in the world is Iguazu? "Results": [ { "TypeName": "LOC", "Text": "Iguazu", "Start": 22, "End": 28 } ] Considering the input sentence below, what is the output result? Widely considered to be one of the most spectacular waterfalls in the world, the Iguazu Falls on the border of Argentina and Brazil, are a certainly must see attraction in the area.
    \item \textbf{Assistant}: "Results": [ { "TypeName": "LOC", "Text": "Iguazu Falls", "Start": 81, "End": 93 }, { "TypeName": "LOC", "Text": "Argentina", "Start": 111, "End": 120 }, { "TypeName": "LOC", "Text": "Brazil", "Start": 125, "End": 131 } ]
\end{itemize}

\subsection{Inference Details} \label{appendix:inference_details}
We use greedy decoding in all experiments for easily reproducible results.
For the model merging experiments, 
we searched among weights \{0.5, 0.7, 0.9\} for the task-specific LoRA modules on the \textsc{MASSIVE} development set and chose 0.9 for our experiments.

\subsection{Details for Retrieval}
To evaluate cross-lingual retrieval performance,
we adapt the implementation from LASER\footnote{\url{https://github.com/facebookresearch/LASER/tree/main/tasks/xsim}} \cite{schwenk-etal-2021-wikimatrix}
to process representations extracted offline.

\section{Results for Individual Languages}
\label{sec:appendix_individual_languages}
The detailed results for \autoref{tab:overall_results} are in \autoref{tab:full_results_slot_filling} (slot filling) and \autoref{tab:full_results_machine_translation} (machine translation).

\section{Results of Aligning at Several Layers}
\label{sec:appendix_align_several_layers}
\begin{table}[t!]
    \small
    \centering
    \begin{tabular}{l c c c c c c ccc}
    \toprule
    &\textbf{Supervised{\tiny $\uparrow$}} 
    & \textbf{Transfer{\tiny $\uparrow$}}
    \\
    \midrule
    \multicolumn{3}{l}{
    \textbf{Slot filling} (\textsc{MASSIVE}): F$_1$}  
    \\
    SFT baseline &
    76.6 &
    60.2 
    \\
    Middle (layer 16) &
    77.0 & 
    61.7 
    \\
    Top (layer 32) &
    76.6 & 
    62.0 
    \\
    Bottom (layer 8) &
    76.8 & 
    58.0 
    \\
    Middle + Bottom &
    77.6 &
    62.5
    \\
    \midrule
    \multicolumn{3}{l}{
    \textbf{Machine translation} (\textsc{WMT}23): COMET
    } 
    \\
    SFT baseline &
    81.5 &
    79.6 
    \\
    Middle (layer 16) &
    81.5 & 
    80.7 
    \\
    Top (layer 32) &
    82.0 & 
    80.2 
    \\
    Bottom (layer 8) &
    81.2 & 
    80.1 
    \\
    Middle + Bottom &
    81.5 &
    80.6
    \\
    \bottomrule
    \end{tabular}
    \caption{Impact of alignment loss placement on supervised and transfer performance on Llama 3. 
    \label{tab:loss_position_combined}}
\end{table}

In \autoref{tab:loss_position_combined}, we show that adding alignment losses at both middle and top layers brings further improvements on slot filling, 
but does not on machine translation.
This task-dependent behavior indicates that how to best align multiple layers still requires further investigation.

\end{document}